\documentclass[letterpaper, 10 pt, conference]{ieeeconf}  

\usepackage{times}
\usepackage{epsfig}
\usepackage{graphicx}
\usepackage{amsmath}
\usepackage{amssymb}

\usepackage{tikz}
\usepackage{comment}
\usepackage[utf8]{inputenc} 
\usepackage[T1]{fontenc}    
\usepackage{url}            
\usepackage{amsfonts}       
\usepackage{nicefrac}       
\usepackage{microtype}      
\usepackage[bb=boondox]{mathalfa}
\usepackage{multirow}
\usepackage{algorithmic}
\usepackage{algorithm}
\usepackage{subfigure}

\usepackage{booktabs}       

\usepackage{float}
\usepackage{caption}
\newcommand{\name}{MANTM}
\newcommand{\planner}{{HTP}}
\usepackage[capitalize]{cleveref}
\crefname{section}{Sec.}{Secs.}
\Crefname{section}{Section}{Sections}
\Crefname{table}{Table}{Tables}
\crefname{table}{Tab.}{Tabs.}

\IEEEoverridecommandlockouts                              

\title{\LARGE \bf
Active Neural Topological Mapping for Multi-Agent Exploration
}

\author{Xinyi Yang$^{1*}$, Yuxiang Yang$^{1*}$, Chao Yu$^{1{\dag}}$, Jiayu Chen$^{1}$, Jingchen Yu$^{1}$, Haibing Ren$^{2}$, \\Huazhong Yang$^{1}$ and Yu Wang$^{1{\dag}}$  
\thanks{*Equal contribution}
\thanks{{\dag}Corresponding authors}
\thanks{$^{1}$Department of Electronic Engineering, Tsinghua University, Beijing, 100084, China.
        {\tt\small yang-xy20@mails.tsinghua.edu.cn; {yanghz, yu-wang}@tsinghua.edu.cn}}%
\thanks{$^{2}$Meituan, 7 Rongda Road, Chaoyang District, Beijing, 100012, China.
        }%
\thanks{Website: \protect\url{https://sites.google.com/view/mantm}}
\thanks{{This research was supported by National Natural Science Foundation of China (No.62203257, 62325405), Tsinghua University Initiative Scientific Research Program, Tsinghua-Meituan Joint Institute for Digital Life, Beijing National Research Center for Information Science, Technology (BNRist) and Beijing Innovation Center for Future Chips.}
}}

\begin{document}

\maketitle
\thispagestyle{empty}
\pagestyle{empty}


\begin{abstract}
This paper investigates the multi-agent cooperative exploration problem, which requires multiple agents to explore an unseen environment via sensory signals in a limited time. 
A popular approach to exploration tasks is to combine active mapping with planning. Metric maps capture the details of the spatial representation, but are with high communication traffic and may vary significantly between scenarios, resulting in inferior generalization. Topological maps are a promising alternative as they consist only of nodes and edges with abstract but essential information and are less influenced by the scene structures. However, most existing topology-based exploration tasks utilize classical methods for planning, which are time-consuming and sub-optimal due to their handcrafted design. Deep reinforcement learning (DRL) has shown great potential for learning (near) optimal policies through fast end-to-end inference. In this paper, we propose \underline{M}ulti-\underline{A}gent \underline{N}eural \underline{T}opological \underline{M}apping (\name) to improve exploration efficiency and generalization for multi-agent exploration tasks. {\name} mainly comprises a Topological Mapper and a novel RL-based \underline{H}ierarchical \underline{T}opological \underline{P}lanner (\planner). The Topological Mapper employs a visual encoder and distance-based heuristics to construct a graph containing main nodes and their corresponding ghost nodes. The {\planner} leverages graph neural networks to capture correlations between agents and graph nodes in a coarse-to-fine manner for effective global goal selection. Extensive experiments conducted in a physically-realistic simulator, Habitat, demonstrate that {\name} reduces the steps by at least 26.40\% over planning-based baselines and by at least 7.63\% over RL-based competitors in unseen scenarios.\looseness=-1



\end{abstract}


\section{Introduction}
\label{sec:intro}

Exploration~\cite{ramakrishnan2021exploration} is one of the fundamental building blocks for developing intelligent autonomous agents, and is widely applied in autonomous driving~\cite{autonomousdriving}, disaster rescue~\cite{rescue}, and planetary exploration~\cite{DBLP:journals/corr/abs-2002-00515}. 
In this paper, we focus on multi-agent exploration, where agents simultaneously explore an unknown scene via sensory signals. To achieve efficient exploration, we typically employ a workflow of autonomous map construction and collaborative planning. 

The spatial arrangement of metric maps~\cite{ans,singleagent-RL1} can vary significantly between scenes, which hinders the generalization ability for exploration. Metric maps also perform poorly in efficiency, as they are with high communication traffic and have difficulty scaling to larger environments. In contrast, topological maps, which contain abstract but essential information, require less communication traffic and are less sensitive to changes in scene structure. Therefore, applying topological maps~\cite{vgm,topo-map2} offers significant generalization potential and high efficiency.

Topology-based exploration tasks commonly utilize classical methods~\cite{normalized_frontier,zhang2022mr} for planning due to their minimal training time and direct deployment for evaluation. However, they often suffer from numerous handcrafted parameters and rule-based coordination strategies, which limits their effectiveness. In contrast, DRL has shown potential for topological exploration~\cite{topo_exploration, topo_multi2} due to its ability to model arbitrarily complex strategies and execute real-time actions. However, these methods are based on pre-built graphs~\cite{topo_exploration,topo-map4} or tested on simple grid maps~\cite{topo_multi2}. 
Applying active topological mapping in RL-based multi-agent exploration is confronted with the following limitations: (a) the number of nodes in the merged graph is large and constantly changing during exploration, leading to unstable RL training and suboptimal results in such a large and varying search space; (b) capturing complex relationships between agents and topological maps is difficult, resulting in an unbalanced workload distribution among agents.






To address the above challenges, we propose \emph{\underline{M}ulti-\underline{A}gent \underline{N}eural \underline{T}opological \underline{M}apping} (\name), an RL-based topological solution for multi-agent exploration. We adopt a modular exploration strategy that divides the planning process into two phases. In the first phase, the global planner infers global goals in each global decision-making step. Subsequently, the local planner predicts the environmental actions to encourage the agents to reach the global goals. {\name} comprises a Topological Mapper to build topological maps, a Hierarchical Topological Planner (\planner) to infer the global goals, and a Local Planner and a Local Policy to generate environmental actions. The Topological Mapper employs a visual encoder and distance-based heuristics to construct graphs with main nodes (i.e., explored areas) and ghost nodes (i.e., unexplored areas). To build more accurate graphs, each agent also maintains a predicted metric map to identify explored areas and prune graphs. We remark that the metric maps are not shared among agents and are not used for global planning, thus promising reduced communication traffic and enhanced generalization capabilities. The RL-based global planner, the Hierarchical Topological Planner ({\planner}), is the most crucial component that selects a main node and then chooses a corresponding ghost node as a global goal for each agent at each global step. This hierarchical goal selection significantly reduces the search space with much fewer candidate nodes, thus addressing challenge (a).
Furthermore, the {\planner} leverages graph neural networks to capture the relationship between agents and topological maps, solving challenge (b).


We conduct extensive experiments in the physically realistic simulator, Habitat~\cite{habitat}. The experimental results demonstrate that {\name} has at least 7.63\% fewer steps than RL-based baselines, and at least 26.40\% fewer steps than planning-based competitors in unseen scenarios.


\section{Related Work}

\subsection{Multi-Agent Exploration} 
In classical visual exploration, agents first perform simultaneous localization and mapping (SLAM)~\cite{SLAM1} to locate their position and reconstruct 2D maps from sensory signals. They then utilize search-based planning algorithms to generate valid exploration trajectories. Representative works mainly include frontier-based methods~\cite{RRT,CoScan}, sampling-based methods~\cite{sample1,sample2}, and graph-based methods~\cite{normalized_frontier,Voronoi}. However, these solutions suffer from expensive computational costs and limited representation capabilities. Recently, deep reinforcement learning~\cite{cvpr22,RL_multi2, yang2023learning} has attracted significant attention due to its powerful expressiveness. 
NeuralCoMapping~\cite{cvpr22} utilizes a multiplex graph neural network to choose effective frontiers as global goals. MAANS~\cite{RL_multi2} uses a transformer-based architecture to infer spatial relationships and intra-agent interactions. However, all these approaches are based on metric maps, which are sensitive to different scene structures and result in subpar generalization. In this paper, we introduce an RL-based topological approach for efficient exploration and superior generalization in unseen scenarios.\looseness=-1


\subsection{Spatial Representation}
Spatial representation usually includes two main types of maps: metric and topological maps. Metric maps are grid maps where each grid predicts its traversability~\cite{ans,save}. However, metric maps struggle with generalization due to significant structural variations across different scenes.
In contrast, topological maps~\cite{graph_generalization} abstractly preserve essential environmental features with nodes and edges, offering a potential solution for improved generalization.
Several works~\cite{topo_exploration,topo-map4} are based on pre-built graphs, focusing on graph refinement or finding optimal paths for coverage.
Recent literature~\cite{vgm,topo-map2,norl} utilizes active topological mapping for navigation tasks. For instance, \cite{vgm} employs the cosine similarity of visual embeddings to construct graphs. However, it requires expert trajectories which are difficult to acquire in the NP-hard multi-agent exploration problem.
Furthermore, \cite{norl} leverages depth images to predict explored/unexplored nodes, while \cite{topo-map2} utilizes semantics for approximate geometric reasoning in topological representations.
However, \cite{norl,topo-map2} requires a predefined goal to predict the geodesic distance from unexplored nodes and select a node with the shortest predicted distance. This is unsuitable for multi-agent exploration where there are no pre-defined goals. 
In this work, we introduce an RL-based Hierarchical Topological Planner to effectively apply active topological mapping in multi-agent exploration. \looseness=-1



\begin{figure*}[t!]
	\centering
 \vspace{2mm}
    \includegraphics[width=0.96\linewidth]{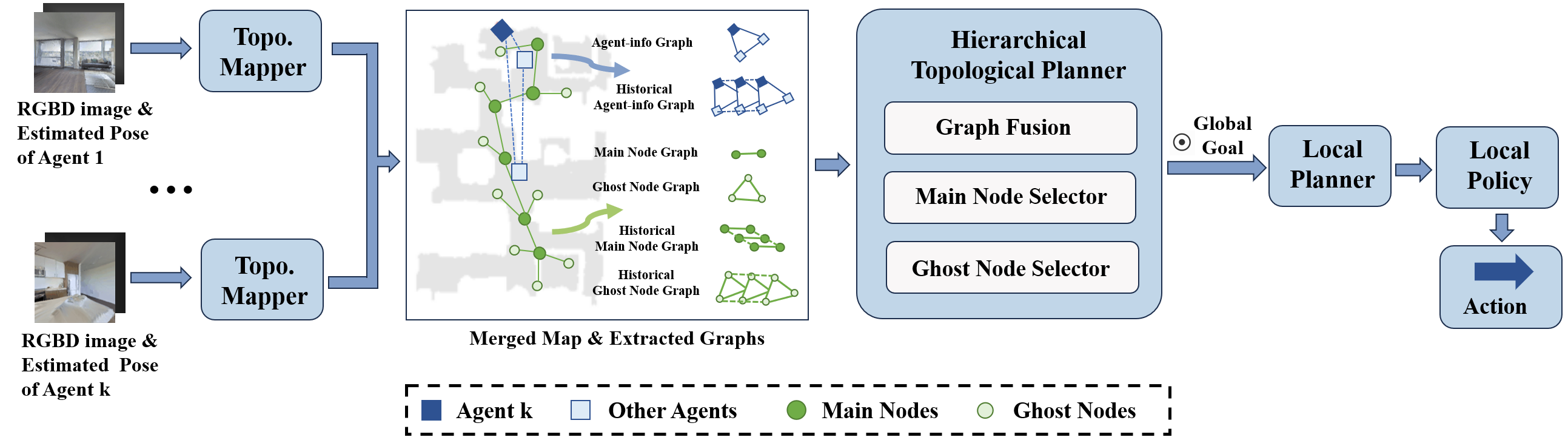}
	\centering \caption{Overview of \emph{Multi-Agent Neural Topological Mapping} ({\name}). Here, we take Agent $k$ as an example.}
\label{fig:Overview}
\vspace{-5mm}
\end{figure*}
\subsection{Graph Neural Networks}
Graph neural networks (GNNs)~\cite{gnn} are widely utilized in multi-agent systems to model interactions between agents. 
\cite{gnn_hierarchical} proposes a hierarchical graph attention network that captures the underlying relationships at the agent and the group level, thereby improving generalization. \cite{vgm} considers GNN as an encoder for mapping tasks to extract node features from images. Additionally, \cite{cvpr22} formulates exploration tasks as bipartite graph matching. In this paper, we propose a hierarchical goal selector based on GNN to capture the correlations between agents and topological maps in a coarse-to-fine manner.\looseness=-1

\section{Task Setup}

Multi-agent cooperative exploration requires agents to explore an unknown scene based on sensory signals.
At each time step, each agent receives a first-person RGB-D image and the estimated pose from
sensors. Agents then perform environmental actions in the physically realistic simulator, Habitat~\cite{habitat}. The horizon of the global decision-making step is 15 steps, and the available environmental actions include \emph{Turn Left}, \emph{Turn Right}, and \emph{Forward}. Following the settings in ANS~\cite{ans} and NRNS~\cite{norl}, we introduce Gaussian noise in the sensor readings and simulate real-world action noise. In the multi-agent scenario, we further consider the following settings. Firstly, we assume perfect communication, where relative spawn locations are shared between agents. This allows us to estimate the relative position of each agent at each timestep by using sensory pose readings and shared spawn locations. Besides, agents are randomly initialized within a 2-meter geodesic distance constraint. This spatially close initialization requires agents to expend more scanning effort for exploration, further increasing the difficulty of cooperation.\looseness=-1



\section{Methodology}
We follow the paradigm of centralized training and decentralized execution (CTDE) with partial observations of $N$ agents, where agent $k$ receives local observation, $o_{t}^k$, at timestep $t$. The overview of {\name} is depicted in \cref{fig:Overview}. Each agent shares the same Topological Mapper, Hierarchical Topological Planner, Local Planner, and Local Policy while making decisions independently. The Topological Mapper of each agent utilizes a visual encoder and distance-based heuristics to construct a topological map based on RGB-D images and estimated poses from sensory signals. For better cooperation, the Topological Mapper transforms all individual maps into the same coordination and merges them. The Hierarchical Topological Planner (\planner) receives graphs that respectively contain current agent information, main node, and ghost nodes, along with corresponding historical graphs containing agent trajectories, selected main nodes, and selected ghost nodes from the merged topological map.
{\planner} utilizes GNN on these graphs to hierarchically capture spatio-temporal and intra-agent information. It then selects a ghost node as a global goal at each global step.
Finally, the Local Planner plans a reachable path to the predicted global goal via the Fast Marching Method (FMM)~\cite{fmm}, and the Local Policy generates environmental actions based on the reachable path. We remark that the Local Planner and the Local Policy do not involve multi-agent interactions, so they are directly adopted from ANS~\cite{ans}.\looseness=-1




\subsection{Topological Mapper\label{sec:mapper}}
We introduce a Topological Mapper to provide a merged topological map, as shown in \cref{fig:Overview}. Inspired by \cite{topo-map2,norl}, we consider two types of nodes in the topological map: main nodes and ghost nodes. 
Main nodes are located where agents have already explored. Ghost nodes are located in the unexplored area and adjacent to the corresponding main node. \looseness=-1

In map construction, the Topological Mapper relies on a visual encoder and distance-based heuristics. Firstly, we adopt an encoder~\cite{vgm} to predict visual embeddings of panoramic RGB-D images. The topological map constructs main nodes based on the cosine similarity between visual embeddings. Specifically, when the cosine similarity between the visual embedding at the current location and that of existing main nodes is below a threshold (i.e., 0.75 in our work), a new main node is constructed at that location and connected to the main node that the agent most recently localized. However, images of some spatially irrelevant locations may be similar (e.g., images of corners and corridors both contain a large portion of the wall), resulting in incorrect connections between main nodes. To deal with this problem, we delete the connection of main nodes with far-distant FMM distance at each global step. 
Once a main node is constructed, $m$ new ghost nodes are uniformly adjacent to this main node at a distance $\lambda$. Ghost nodes can be removed if they are located in the explored areas identified by the current agent's predicted metric map via SLAM. We remark that the metric map is just for identifying explored regions and not for global planning. Besides, a ghost node can be converted to a main node if the agent passes through it. Therefore, the potential number of remaining ghost nodes ranges from 0 to $m$$\times$$N_{main}$, where $N_{main}$ is the number of main nodes. In our work, we choose $\lambda=3$ meters and $m=12$. Moreover, we delete spurious ghost nodes and their connected edges based on the FMM distance between two ghost nodes belonging to two different main nodes and between a main node and a ghost node belonging to another main node.\looseness=-1

For better cooperation, we transform all individual topological maps into the same coordinate system and merge them based on the estimated poses of the agents at each global step. During merging, if the distance between two main nodes from different maps is below a threshold (i.e., 3 meters in our work), we randomly remove one of the nodes and redirect its connected edges to the remaining node.

\begin{figure*}[ht]
	\centering
    \includegraphics[width=0.95\linewidth]{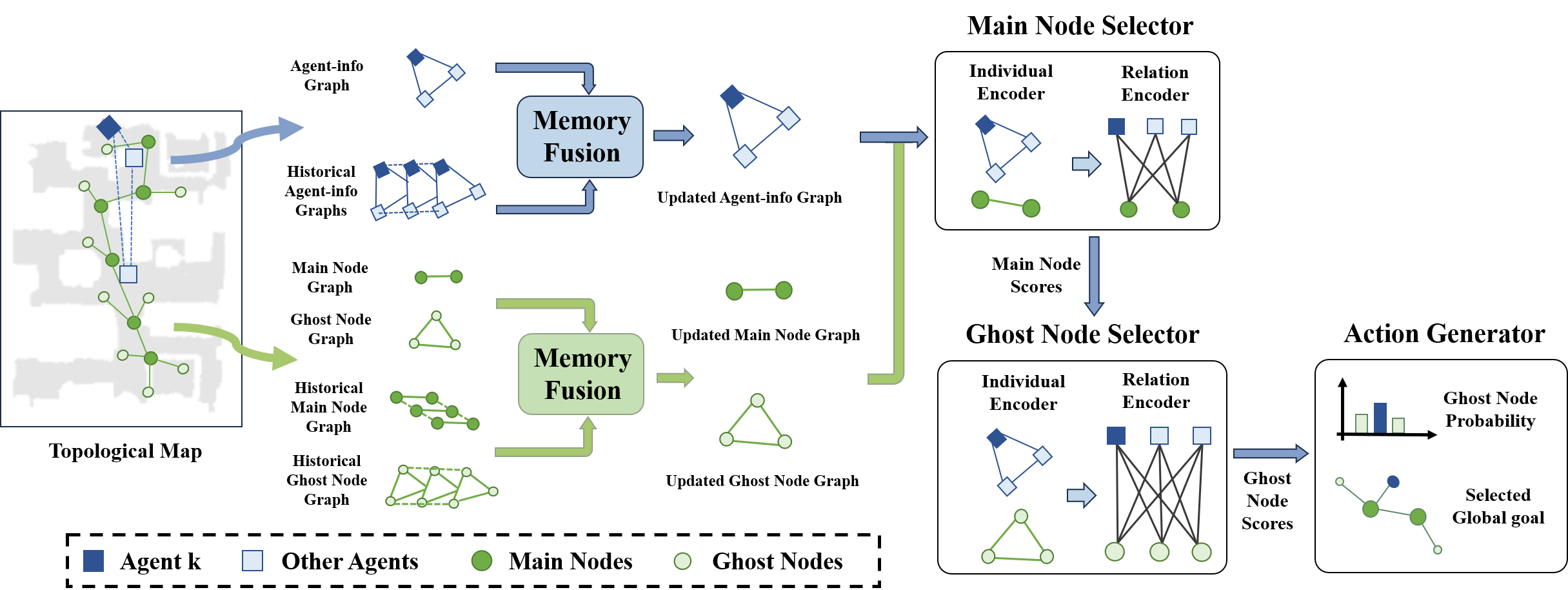}
	\centering \caption{Workflow of \emph{Hierarchical Topological Planner} (\planner), including a Memory Fusion, a Main Node Selector, a Ghost Node Selector and an Action Generator. {\planner} is under decentralized decision-making setting. We take Agent $k$ as an example.}
\label{fig:htp}
\vspace{-6mm}
\end{figure*}
\subsection{Hierarchical Topological Planner}
The Hierarchical Topological Planner (\planner) selects a ghost node as a global goal at each global step. However, it is difficult for MARL to directly explore the (near) optimal strategy due to the large and varying search space associated with the number of ghost nodes. To address this issue, we propose a hierarchical network that matches agents to main nodes and then to the corresponding ghost nodes. Therefore, the {\planner} can be easily optimized in the reduced search space with much fewer candidate nodes and potentially provide a more appropriate probability distribution of ghost nodes in a coarse-to-fine manner.
The workflow of the {\planner} is illustrated in \cref{fig:htp}.
Firstly, at each global step, we extract an agent-info graph, $G_{a}$, a main node graph, $G_{m}$, a ghost node graph, $G_{g}$, and three corresponding historical graphs, $\hat{G}$, from a merged topological map. Subsequently, we update each $G$ with its historical graph in the Memory Fusion to aggregate current and historical information. The Main Node Selector computes matching scores between the agent and the main nodes via the attention mechanism~\cite{attention}. 
Afterward, the Ghost Node Selector calculates the probability distribution of ghost nodes based on ghost node features and the matching results from the Main Node Selector. 
Finally, the Action Generator takes in the probability distribution of ghost nodes and determines the most appropriate ghost node as the global goal at each global decision-making step.\looseness=-1

We remark that $G_{a}$ contains the current agent information, while $G_{m}$ contains current main node features, and $G_{g}$ includes current ghost node features. Besides, in historical graphs, $\hat{G_{a}}$ contains agent trajectories, $\hat{G_{m}}$ contains selected main node features, and $\hat{G_{g}}$ includes selected ghost node features from past global steps.
Each node consists of its grid position, $(x,y)\in R^2, x,y\in [0,map\ size]$, and a semantic label, $(s_1,s_2)\in N^2, N=\left\{0,1\right\}$.
This semantic label represents the type of node, including agent nodes, historical agent nodes, map nodes, and historical map nodes.

\subsubsection{Memory Fusion}
Exploration is a memory-based task~\cite{memory_based} which heavily relies on historical information.
Therefore, this module incorporates each graph with its historical counterpart to mitigate the risk of memory loss. Concretely, we first use a weight-shared multi-layer perception (MLP) layer to encode node features due to their consistent information types across different graphs. Then, the Memory Fusion merges each graph with its corresponding historical graph and yields an updated graph, $\tilde{G}$, via the self-attention and the cross-attention mechanisms~\cite{attention}. The output dimensions of the last cross-attention layer match those of the original node features, ensuring that the shape of $\tilde{G}$ remains consistent with that of $G$.\looseness=-1

\subsubsection{Main Node Selector}
The Main Node Selector is introduced for the high-level goal selection and yields the matching scores between the agent and the main nodes. It predicts the next preferred region to explore since a main node with a higher matching score implies a higher probability of selecting its corresponding ghost nodes as the global goal. More concretely, this module receives $\tilde{G}_{a}$ and $\tilde{G}_{m}$ and then leverages an Individual Encoder and a Relation Encoder to infer the matching scores, $S_{m, re}$. The Individual Encoder perceives the states of the agents and the spatial information of the main nodes, while the Relation Encoder captures the correlation between the agents and the main nodes. \looseness=-1

\begin{figure}[ht]
\vspace{-1mm}
\captionsetup{justification=centering}
	\centering
    \subfigure[\label{fig:ind_encoder}Individual Encoder]
        {\centering        {\includegraphics[height=4.5cm]{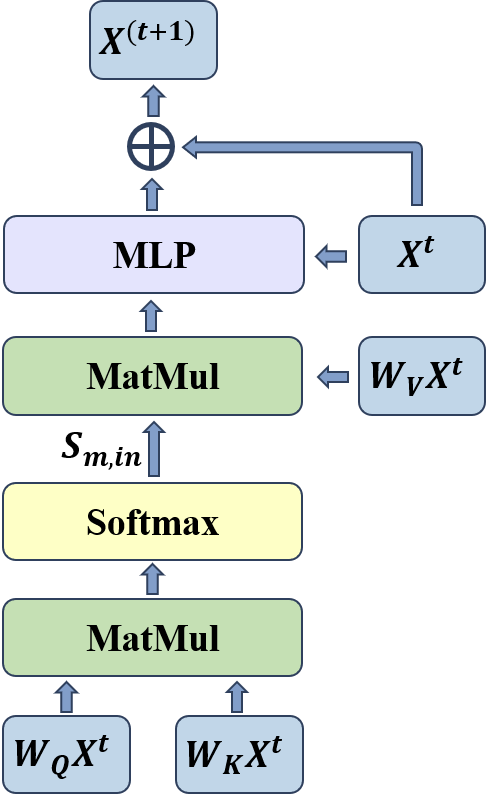}
            }
    }
    \subfigure[\label{fig:rel_encoder}Relation Encoder]
        {\centering        {\includegraphics[height=4.5cm]{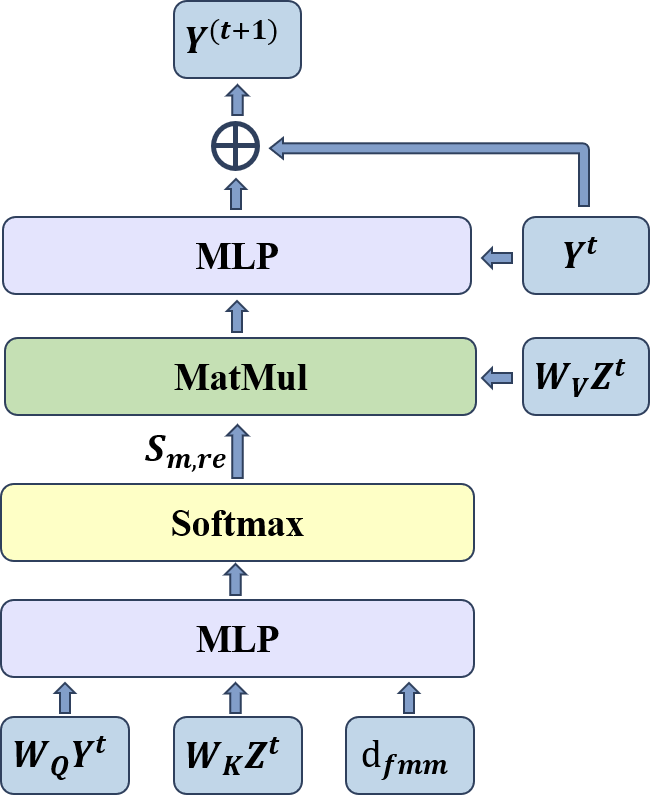}
            }
    }
    \vspace{-2mm}
    \caption{\raggedright{Network architecture of the Individual Encoder and the Relation Encoder.}}
    \vspace{-3mm}
    \label{fig:encoder}
\end{figure}

\textbf{Individual Encoder:} The Individual Encoder captures relationships between any two nodes in the same graph and updates these nodes. As shown in \cref{fig:ind_encoder}, this module first calculates the normalized matching scores of any two nodes:
\begin{equation}
S_{m, in} = Softmax(W_{Q}X \times W_{K}X).
\end{equation}
Here $X$ denotes the node features, while $W_{Q}$ and $W_{K}$ represent the linear projections of $X$. The $S_{m, in}$ is a matrix where each element, $S_{m, in}^{(i,j)}$, refers to the matching score between node $i$ and node $j$.

After that, each node is updated by an MLP layer, $f_{in}$, with a residual connection. The input to $f_{in}$ is the concatenation of each node feature and the weighted sum of its neighboring features. The update of each node is formulated as: \looseness=-1
\begin{equation}
    \begin{aligned}
        X^{(t+1)} = X^t
        +f_{in}(X^t,  W_{V}X^t \times S_{m, in}^T),
    \end{aligned}
\end{equation}
where $X^{t}$ represents the node features at time step $t$, and $W_{V}$ is the linear projection of $X^t$. The matching scores between nodes in a graph, $S_{m, in}$, serve as the weights for neighboring nodes. 

\textbf{Relation Encoder:} The Relation Encoder captures correlations between any two nodes from different graphs. The architecture of the Relation Encoder is shown in \cref{fig:rel_encoder}. Considering the node features in two different graphs as $Y$ and $Z$, we calculate the normalized matching scores for each node pair $(y,z)$ by taking a softmax operator over the outputs of an MLP layer, $f_{dis}$. The input to $f_{dis}$ is the concatenation of $W_{Q}Y$, $W_{K}Z$, and $d_{fmm}$, where $d_{fmm}$ is the FMM distance between the given node pair. The calculation of the scores is formulated as:
\begin{equation}
    S_{m, re} = Softmax( f_{dis}( W_{Q}Y,
    W_{K}Z, d_{fmm})).
\end{equation}
Afterward, we update all nodes via an MLP layer, $f_{re}$, with a residual connection:
\begin{equation}
Y^{(t+1)} =  Y^t + f_{re}(Y^t,  W_{V}Z^t \times S_{m, re}^T).
\end{equation}
Here $W$ represents linear projections, while $Y^t$ and $Z^t$ denote different node features at time step $t$. Note that $S_{m, re}^{(k)}$ denotes the matching score between agent $k$ and the main nodes and is then sent to the Ghost Node Selector for agent $k$.

\subsubsection{Ghost Node Selector}
For the low-level goal selection, the Ghost Node Selector provides a probability distribution of ghost nodes and predicts the next location to explore based on the preferred region. Similar to the Main Node Selector, the Ghost Node Selector first receives $\tilde{G}_{a}$ and $\tilde{G}_{g}$ and provides ghost node scores, $S_{g, re}$, via an Individual Encoder and a Relation Encoder. The Individual Encoder captures intra-agent interactions and spatial information of ghost nodes, while the Relation Encoder infers the correlations between agents and ghost nodes. Additionally, we update $S_{g, re}^{(k,j)}$ by multiplying it with $S_{m, re}^{(k,h)}$ to aggregate information from main and ghost nodes for more appropriate goal selection:
\begin{equation}
    \hat{S}_{g, re}^{(k,j)} =  S_{g, re}^{(k,j)} \times S_{m, re}^{(k,h)}.
\label{eq:multiply}
\end{equation}
Here, $k$ is a specific agent, $j$ is the main node, and $h$ denotes the corresponding ghost node of the main node $j$. $\hat{S}_{g, re}^{(k)}$ is the final matching score of agent $k$ and ghost nodes.

\subsubsection{Action Generator}
The Action Generator selects a global goal from all ghost nodes at each global step. We regard $\hat{S}_{g, re}$ as the probability distribution over the ghost nodes. The action space for the {\planner} is $A=\{a|a\in ghost\ nodes\}$, where $a$ is a discrete variable sampled from a categorical distribution, indicating the index of the selected ghost node.



\subsection{Reinforcement Learning Design}
We train the {\planner} by using multi-agent proximal policy optimization (MAPPO)~\cite{mappo}, which is a multi-agent variant of proximal policy optimization (PPO)~\cite{ppo}.

\textbf{Reward Function: }To promote efficient and cooperative exploration, we introduce four types of rewards. The coverage reward, $R_{cov}$, represents the increase of the explored area to encourage thorough exploration. Besides, the success reward, $R_{suc}$, offers a bonus for reaching the target coverage ratio. The overlap penalty, $R_{o}$, denotes the overlapped area passed by agents to encourage cooperation. Finally, for efficient exploration, agents receive a constant time penalty, $R_{t}$, at each timestep until they attain the target coverage ratio. As a result, the reward function, $R_{total}$, is a linear combination of these kinds of rewards.


\section{Experiments}

\subsection{Experimental Details}
We conduct all experiments in the Habitat simulator~\cite{habitat}, using the Gibson Challenge dataset~\cite{dataset1} and the Habitat-Matterport 3D dataset (HM3D)~\cite{dataset2}. The scenes in these datasets are collected from real building-scale residential, commercial, and civic spaces using 3D scanning and reconstruction. We filter out some scenes that are inappropriate for our task, following \cite{RL_multi2}, which is one of the best RL-based approaches for cooperative exploration. This filtering process involved removing scenes with large disconnected regions or multiple floors where agents couldn't attain 90\% coverage of the entire house. 
Furthermore, we exclude scenes smaller than 70 $m^2$, as their topological maps would contain too few nodes to fully show the advantages of graphs.
To better demonstrate the robustness of {\name} on training scenes and its effective generalization to novel scenes, we follow \cite{RL_multi2,cvpr22} by dividing the remaining scenes into $10$ training scenes from the Gibson Challenge dataset and $28$ testing scenes from both datasets.
We perform RL training with $10^6$ timesteps over 3 random seeds. Each evaluation score has the format of ``mean (standard deviation)'', and is averaged over 300 testing episodes.\looseness=-1





\begin{table*}
\vspace{2mm}
\centering
\footnotesize
\scalebox{1.}{
\footnotesize
\setlength\tabcolsep{1.5mm}{\begin{tabular}{ccrccc|ccc|c} 
\toprule
Agents     &Sce.                                                                                    & Metrics                    &CoScan  &NF& Voronoi & ANS-Merge &NCM& MAANS & {\name}  \\ 

\midrule
\multirow{6}{*}{N =3} & 
\multirow{3}{*}{\begin{tabular}[c]{@{}c@{}}Middle\\(\textgreater{}$70m^2$)\end{tabular}} & \textit{Mut. Over.} $\downarrow$   &  0.39\scriptsize{(0.01)}   &  0.56\scriptsize{(0.01)}  & \textbf{0.29\scriptsize{(0.01)}}       &   0.44\scriptsize{(0.01)} & 0.38\scriptsize{(0.01)} & 0.39\scriptsize{(0.01)}  & 0.36\scriptsize{(0.02) } \\ 
\cmidrule{3-10}
                       & & \textit{Steps} $\downarrow$    &     244.86\scriptsize{(7.25)} &    231.72\scriptsize{(5.43)}  &     226.35\scriptsize{(3.51)}  
                             &                    211.88\scriptsize{(4.23)} &202.20\scriptsize{(5.42)}  &  185.31\scriptsize{(5.81)} & \textbf{166.59\scriptsize{(4.19)} }   \\ 
\cmidrule{3-10}
                     & & \textit{Coverage} $\uparrow$   &0.89\scriptsize{(0.02) }     &  0.92\scriptsize{(0.01)}& 0.92\scriptsize{(0.01)}    & 
                     0.94\scriptsize{(0.01)} &  \textbf{0.97\scriptsize{(0.01)}}& 0.96\scriptsize{(0.01)} &   \textbf{0.97\scriptsize{(0.01)}}             \\ 
\cmidrule{2-10}
&\multirow{3}{*}{\begin{tabular}[c]{@{}c@{}}Large\\(\textgreater{}$100m^2$)\end{tabular}} & \textit{Mut. Over.}$\downarrow$ &  0.42\scriptsize{(0.01)}  & 0.65\scriptsize{(0.03)} & \textbf{0.33\scriptsize{(0.01)}}    &   0.52\scriptsize{(0.02)} &0.45\scriptsize{(0.01)} &   0.43\scriptsize{(0.01)} &      0.38\scriptsize{(0.01) }       \\ 
\cmidrule{3-10}
                     & & \textit{Steps} $\downarrow$      &485.74\scriptsize{(6.89)} &454.34\scriptsize{(6.89)}& 439.81\scriptsize{(8.47)}     & 
                  386.54\scriptsize{(11.96)} &381.84\scriptsize{(8.63)}& 379.45\scriptsize{(5.69)} & \textbf{323.72\scriptsize{(10.64)}}    \\ 
\cmidrule{3-10}
                  & & \textit{Coverage} $\uparrow$   
                  &0.90\scriptsize{(0.03) } & 0.92\scriptsize{(0.01)} & 0.88\scriptsize{(0.01)}  & 
                    0.94\scriptsize{(0.01)} &  0.93\scriptsize{(0.01)}  & 0.95\scriptsize{(0.01)} &    \textbf{0.96\scriptsize{(0.01)}}        \\
\midrule
\multirow{6}{*}{N = 4} & 
\multirow{3}{*}{\begin{tabular}[c]{@{}c@{}}Middle\\(\textgreater{}$70m^2$)\end{tabular}} & \textit{Mut. Over.} $\downarrow$  &0.36\scriptsize{(0.01)}  & 0.53\scriptsize{(0.03)} & \textbf{0.21\scriptsize{(0.01)}} & 0.42\scriptsize{(0.03)} &0.35\scriptsize{(0.01)} &  0.25\scriptsize{(0.01)}&  0.22\scriptsize{(0.01)}   \\ 
\cmidrule{3-10}
             & & \textit{Steps} $\downarrow$      &  236.81\scriptsize{(4.39)} &219.35\scriptsize{(3.68)}& 218.80\scriptsize{(2.46)}     & 173.30\scriptsize{(4.14)} &174.63\scriptsize{(4.57)}& 162.09\scriptsize{(3.94)}& \textbf{149.72\scriptsize{(1.94)}}  \\ 
\cmidrule{3-10}
                 & & \textit{Coverage} $\uparrow$   & 0.97\scriptsize{(0.03)}   &0.97\scriptsize{(0.01)}& 0.95\scriptsize{(0.01)}     &  \textbf{0.98\scriptsize{(0.01)}} &  0.97\scriptsize{(0.01)}& \textbf{ 0.98\scriptsize{(0.01)}} & \textbf{0.98\scriptsize{(0.01)} }           \\ 
\cmidrule{2-10}
&\multirow{3}{*}{\begin{tabular}[c]{@{}c@{}}Large\\(\textgreater{}$100m^2$)\end{tabular}} & \textit{Mut. Over.}$\downarrow$  & 0.36\scriptsize{(0.01)}  & 0.60\scriptsize{(0.01)}& \textbf{0.25\scriptsize{(0.01)}} & 0.45\scriptsize{(0.03)}& 0.38\scriptsize{(0.01) }&0.38\scriptsize{(0.01) }  &   0.28\scriptsize{(0.01) }    \\ 
\cmidrule{3-10}
                   & & \textit{Steps} $\downarrow$      &479.47\scriptsize{(7.35)} & 425.88\scriptsize{(6.15)}&
                   418.49\scriptsize{(5.23) }       &325.29\scriptsize{(5.48)} &322.27\scriptsize{(8.67)}
                   & 315.80\scriptsize{(3.55)} &\textbf{284.50\scriptsize{(3.66)}}   \\ 
\cmidrule{3-10}
                     & & \textit{Coverage} $\uparrow$     &  0.91\scriptsize{(0.03)} &\textbf{0.96\scriptsize{(0.01)}} 
                &\textbf{0.96\scriptsize{(0.01)}}         & 0.95\scriptsize{(0.01)} &0.95\scriptsize{(0.01)}&   0.95\scriptsize{(0.01)}& \textbf{0.96\scriptsize{(0.01)}}  \\
\bottomrule
\end{tabular}}}
\caption{Performance of {\name}, planning-based baselines, and RL-based baselines with $N=3,4$ agents on the Gibson dataset. Note that the horizon of middle and large maps is 300 steps and 600 steps, respectively.}
\label{tab: gibson_results}
\end{table*}

\begin{table*}
\centering
\footnotesize
\scalebox{.95}{
\setlength\tabcolsep{1.5mm}{\begin{tabular}{ccrccc|ccc|c} 
\toprule
Agents     &Sce.                                                                                    & Metrics                  &CoScan  &NF &Voronoi  & ANS-Merge &NCM&MAANS & {\name}  \\ 
\midrule

\multirow{9}{*}{N =3}& 
\multirow{3}{*}{\begin{tabular}[c]{@{}c@{}}Middle\\(\textgreater{}$70m^2$)\end{tabular}} & \textit{Mut. Over.} $\downarrow$
&0.40\scriptsize{(0.01)} &  0.63\scriptsize{(0.04)}  & \textbf{0.32\scriptsize{(0.03)} }&0.48\scriptsize{(0.03)}   &0.41\scriptsize{(0.03)} &  0.46\scriptsize{(0.07)}   & 0.36\scriptsize{(0.01) }    \\
\cmidrule{3-10}
                  & & \textit{Steps} $\downarrow$   
                  &354.02\scriptsize{(5.40)} &333.09\scriptsize{(4.26)}& 256.78\scriptsize{(2.58)} &279.72\scriptsize{(5.38)}&268.03\scriptsize{(3.43)} & 267.23\scriptsize{(4.82)} & \textbf{238.10\scriptsize{(3.39)}} \\ 
\cmidrule{3-10}
                        & & \textit{Coverage} $\uparrow$    &0.91\scriptsize{(0.01)}&  0.95\scriptsize{(0.01)}&    0.96\scriptsize{(0.01)} & 0.96\scriptsize{(0.01) }&0.96\scriptsize{(0.01) } &   \textbf{0.97\scriptsize{(0.01) }}&    \textbf{0.97\scriptsize{(0.01) }}    \\ 
\cmidrule{2-10}
 &\multirow{3}{*}{\begin{tabular}[c]{@{}c@{}}Large\\(\textgreater{}$100m^2$)\end{tabular}} & \textit{Mut. Over.}$\downarrow$   &  0.40\scriptsize{(0.01)} & 0.63\scriptsize{(0.01)} & \textbf{0.28\scriptsize{(0.03)}}& 0.52\scriptsize{(0.05)}   &0.42\scriptsize{(0.03)}& 0.55\scriptsize{(0.07)} &   0.39\scriptsize{(0.04) }         \\ 
\cmidrule{3-10}
                         & & \textit{Steps} $\downarrow$  & 698.22\scriptsize{(8.16)} &649.60\scriptsize{(11.23)}& 509.24\scriptsize{(5.54)} & 497.41\scriptsize{(11.60)}&463.75\scriptsize{(12.10)} &458.69\scriptsize{(14.04)} & \textbf{419.88\scriptsize{(7.91)} }  \\ 
\cmidrule{3-10}
                         & & \textit{Coverage} $\uparrow$ 
                         &0.82\scriptsize{(0.01)} &0.92\scriptsize{(0.01)}&   0.94\scriptsize{(0.01)} &0.93\scriptsize{(0.01)} &0.95\scriptsize{(0.01)}&    0.95\scriptsize{(0.01)} &  \textbf{ 0.96\scriptsize{(0.01)}}    \\
\cmidrule{2-10}
&\multirow{3}{*}{\begin{tabular}[c]{@{}c@{}}Super Large\\(\textgreater{}$200m^2$)\end{tabular}} & \textit{Mut. Over.}$\downarrow$   &0.33\scriptsize{(0.01)}&0.63\scriptsize{(0.01)} & \textbf{0.28\scriptsize{(0.01)}} &  0.45\scriptsize{(0.04)}&0.41\scriptsize{(0.01)} &   0.40\scriptsize{(0.04)} &   0.35\scriptsize{(0.01)}   \\ 
\cmidrule{3-10}
                          & & \textit{Steps} $\downarrow$   
                          & 1710.88\scriptsize{(41.11)} &1456.80\scriptsize{(48.94)}& 1321.62\scriptsize{(44.73)} &1343.17\scriptsize{(50.68)}&1147.25\scriptsize{(57.26)}& 1135.24\scriptsize{(53.06)} & \textbf{982.54\scriptsize{(43.51)}}   \\ 
\cmidrule{3-10}
                          & & \textit{Coverage} $\uparrow$            &0.82\scriptsize{(0.02)} & 0.87\scriptsize{(0.01)}&    0.87\scriptsize{(0.01)} &0.92\scriptsize{(0.01)}&0.92\scriptsize{(0.01)}&     0.94\scriptsize{(0.02)} &  \textbf{ 0.97\scriptsize{(0.01) } }  \\
\midrule

\multirow{9}{*}{N = 4}& 
\multirow{3}{*}{\begin{tabular}[c]{@{}c@{}}Middle\\(\textgreater{}$70m^2$)\end{tabular}} & \textit{Mut. Over.} $\downarrow$ &0.35\scriptsize{(0.01)} &0.57\scriptsize{(0.02)}  &  \textbf{0.30\scriptsize{(0.02)}} & 0.34\scriptsize{(0.01)} &0.35\scriptsize{(0.03)}&   0.34\scriptsize{(0.01)}   & 0.35\scriptsize{(0.01)} \\ 
\cmidrule{3-10}
                     & & \textit{Steps} $\downarrow$   &280.38\scriptsize{(4.86)} & 314.12\scriptsize{(4.30)} & 233.51\scriptsize{(2.79)} &227.42\scriptsize{(5.81)} & 227.38\scriptsize{(2.10)}&219.68\scriptsize{(5.64)}&\textbf{191.31\scriptsize{(1.66)}}\\ 
\cmidrule{3-10}
                       & & \textit{Coverage} $\uparrow$    &0.95\scriptsize{(0.01)} & 0.96\scriptsize{(0.01)} &  \textbf{0.97\scriptsize{(0.01)}} & \textbf{0.97\scriptsize{(0.01)}}&
                       0.96\scriptsize{(0.02)}&  \textbf{0.97\scriptsize{(0.01)}}& \textbf{0.97\scriptsize{(0.01)} }\\ 
\cmidrule{2-10}
 &\multirow{3}{*}{\begin{tabular}[c]{@{}c@{}}Large\\(\textgreater{}$100m^2$)\end{tabular}} & \textit{Mut. Over.}$\downarrow$  &  0.37\scriptsize{(0.01)} & 0.58\scriptsize{(0.02)} &  \textbf{0.29\scriptsize{(0.01)}} & 0.44\scriptsize{(0.02)}  &0.42\scriptsize{(0.02)}& 0.41\scriptsize{(0.01)}      & 0.38\scriptsize{(0.01)}    \\ 
\cmidrule{3-10}
                 & & \textit{Steps} $\downarrow$  &  608.14\scriptsize{(7.24)} &601.15\scriptsize{(8.60)} & 407.31\scriptsize{(8.34)} &384.27\scriptsize{(9.63)}&389.12\scriptsize{(8.53)}& 363.75\scriptsize{(7.87)} &  \textbf{345.52\scriptsize{(7.85)}}\\ 
\cmidrule{3-10}
                   & & \textit{Coverage} $\uparrow$  &0.92\scriptsize{(0.01)} & 0.90\scriptsize{(0.02)}&   0.96\scriptsize{(0.01)}& 0.95\scriptsize{(0.01)}&
                   \textbf{0.97\scriptsize{(0.01)}} & \textbf{0.97\scriptsize{(0.02)}}& \textbf{0.97\scriptsize{(0.02)}} \\
\cmidrule{2-10}
&\multirow{3}{*}{\begin{tabular}[c]{@{}c@{}}Super Large\\(\textgreater{}$200m^2$)\end{tabular}} & \textit{Mut. Over.}$\downarrow$   & 0.29\scriptsize{(0.01)} &0.59\scriptsize{(0.01)} &   \textbf{0.22\scriptsize{(0.01)}} &0.35\scriptsize{(0.01)} & 0.34\scriptsize{(0.01)} &   0.31\scriptsize{(0.02)}& 0.33\scriptsize{(0.01)} \\ 
\cmidrule{3-10}
                          & & \textit{Steps} $\downarrow$   &1538.13\scriptsize{(22.13)} & 1338.95\scriptsize{(47.36)}&
                          1264.12\scriptsize{(39.34)} & 1156.49\scriptsize{(43.60)}&
                          1064.16\scriptsize{(42.85)}&
                          1012.52\scriptsize{(38.62)}  & \textbf{900.71\scriptsize{(29.24)}} \\ 
\cmidrule{3-10}
                       & & \textit{Coverage} $\uparrow$  &   0.89\scriptsize{(0.01)} & 0.90\scriptsize{(0.01)}&   0.91\scriptsize{(0.02)} &0.91\scriptsize{(0.01)} &0.94\scriptsize{(0.02)}&   0.93\scriptsize{(0.01)}& \textbf{0.97\scriptsize{(0.01)}}    \\
\bottomrule
\end{tabular}}}
\caption{Performance of {\name}, planning-based baselines and RL-based baselines with $N=3,4$ agents on the HM3D dataset. Note that the horizon of middle, large, and super large maps is 450 steps, 720 steps, and 1800 steps, respectively.}
\label{tab: hm3d_results}
\vspace{-6mm}
\end{table*}

\subsection{Evaluation Metrics}
We consider 3 statistical metrics to capture different characteristics of a particular exploration strategy. These metrics are only for analysis, and we primarily focus on \emph{Steps} as our performance criterion.

\begin{itemize}
\setlength{\parskip}{0pt} \setlength{\itemsep}{0pt plus 1pt}
    \item \textbf{Steps:  }This metric considers the timesteps required to achieve 90\% coverage within an episode. Fewer \emph{Steps} imply faster exploration.
    \item \textbf{Coverage:  }This metric denotes the final ratio of the explored area to total explorable area at the end of the episode. A higher \emph{Coverage} ratio reflects a more effective exploration.
    \item \textbf{Mutual Overlap: }This metric shows the ratio of the overlapped area to the currently explored area when the \emph{Coverage} ratio achieves 90\%. Lower \emph{Mutual Overlap} ratio indicates better collaboration.
\end{itemize}
\vspace{-\topsep}


\subsection{Baselines}
We challenge {\name} against three representative planning-based approaches (CoScan, Topological Frontier, Voronoi) and three prominent RL-based solutions (ANS-Merge, NeuralCoMapping, MAANS). Note that Topological Frontier and Voronoi are also graph-based approaches.


\begin{itemize}
\setlength{\parskip}{0pt} \setlength{\itemsep}{0pt plus 1pt}
 \item \textbf{CoScan}~\cite{CoScan}: This frontier-based method applies k-means clustering to all frontiers and assigns a frontier cluster to each agent. Afterward, each agent plans an optimal traverse path over the assigned frontiers.
 \item \textbf{Topological Frontier (TF)}~\cite{normalized_frontier}: This graph-based approach calculates a normalized traveling cost for each ghost node built from the Topological Mapper and considers the node with the lowest cost as the global goal.\looseness=-1
 \item \textbf{Voronoi}~\cite{Voronoi}: This graph-based solution divides the map into several parts and transforms it into a Voronoi graph. Each agent then only searches the unexplored region in its partition, reducing the overlapped area.
 \item \textbf{ANS-Merge}~\cite{ans}: ANS is exemplary in RL-based single-agent exploration. It takes in egocentric local and global metric maps and infers global goals for the agents. We extend ANS to multi-agent exploration by sending merged maps to the global planner and use the same reward function as ours.
 \item\textbf{NeuralCoMapping (NCM)}~\cite{cvpr22}: NeuralCoMapping introduces a multiplex graph neural network to predict the neural distance between frontier nodes and agents. It then assigns each agent a frontier node based on the neural distance in each global step.
 \item \textbf{MAANS}~\cite{RL_multi2}: MAANS is a variant of ANS for multi-agent exploration. This method leverages a transformer-based Spatial-TeamFormer to enhance cooperation. For a fair comparison, we conduct training on 10 maps without the policy distillation mentioned in \cite{RL_multi2}.
\end{itemize}

We remark that {\name} and the baselines are under the same assumptions in our task. All the baselines only replace the Topological Mapper and the {\planner} with alternatives and keep the rest the same as {\name}, except for TF, which only substitutes the {\planner}.


\begin{figure*}[ht]
\captionsetup{justification=centering}
	\centering
    \subfigure[\label{fig:casea}Comparison of Map Construction]
        {\centering
        {\includegraphics[height=4cm]{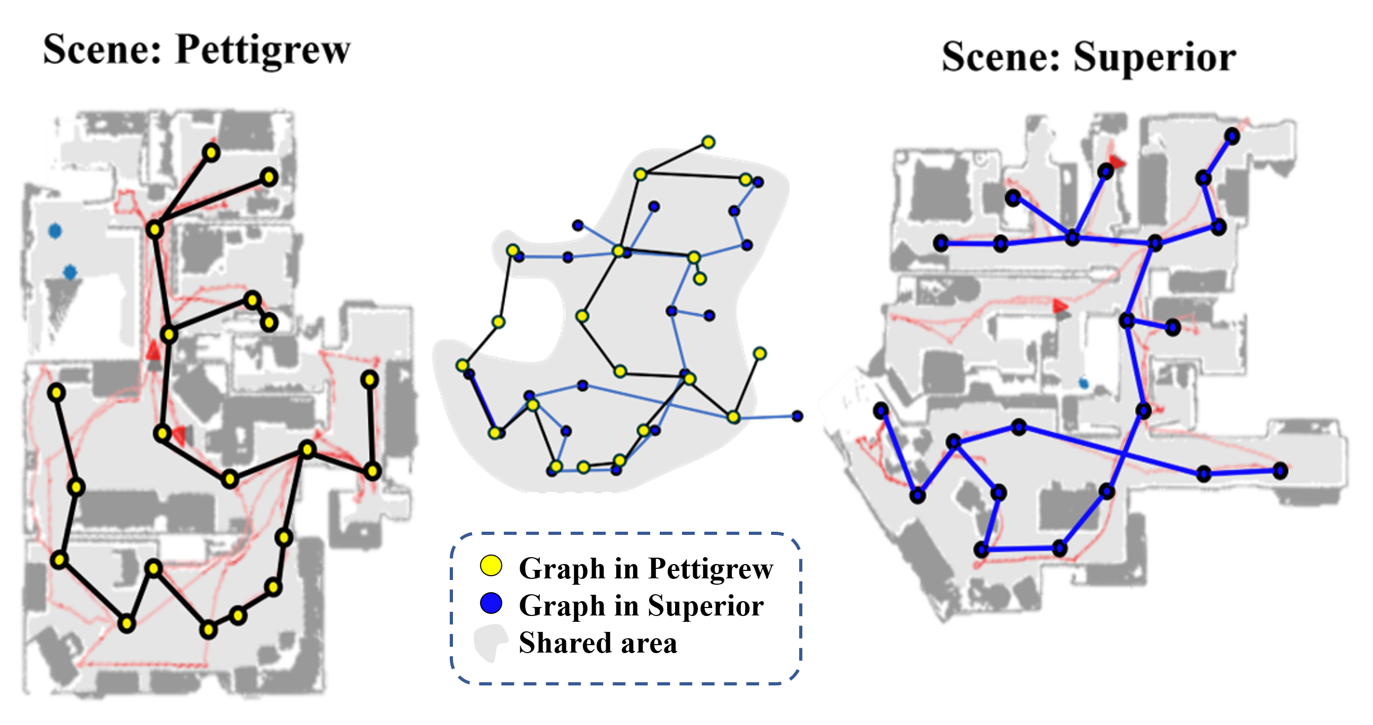}
            }
    }
    \subfigure[\label{fig:caseb}Comparison of Planning Strategy]
        {\centering
        {\includegraphics[height=4cm]{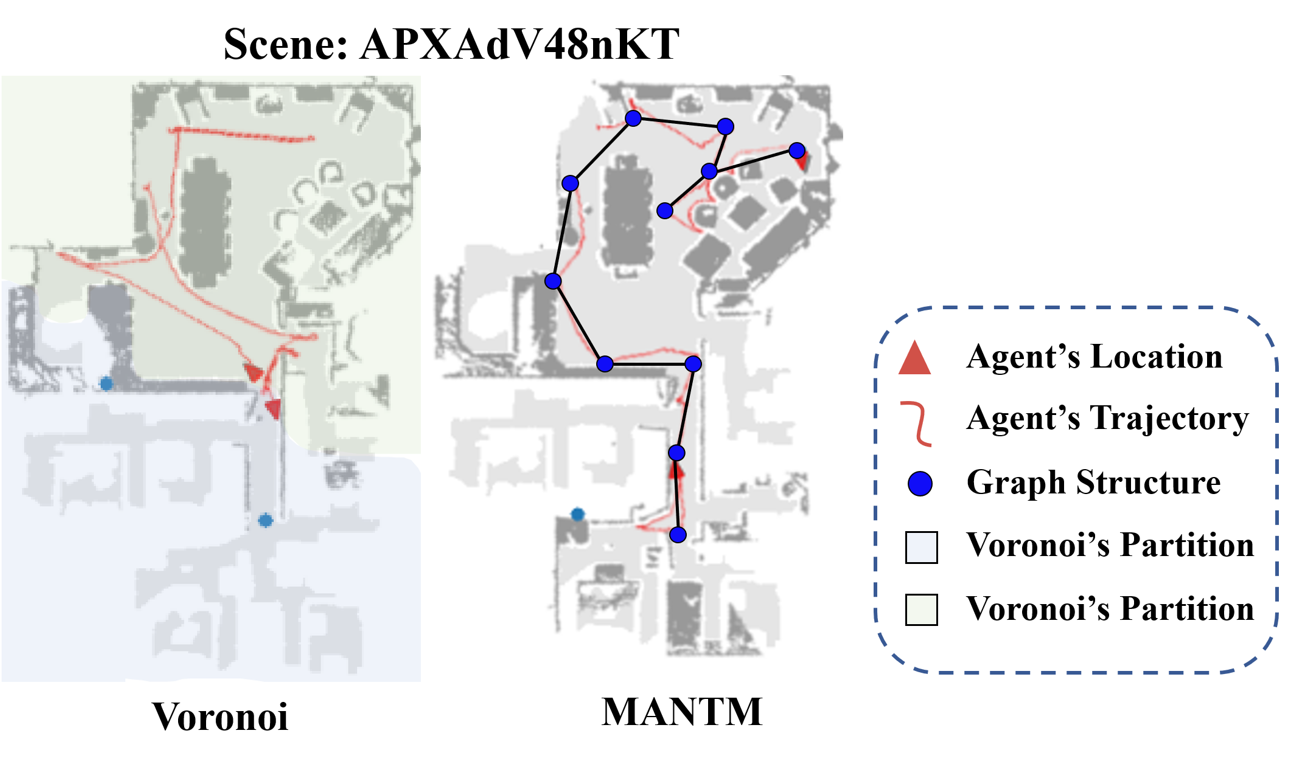}
    	}
    }
     \vspace{-2mm}
    \caption{\raggedright{Case studies of Map Construction and Planning Strategy. (a) shows that the graph structures of different scenes seem congruous in general, marked in grey. (b) displays agents' trajectories in Voronoi and {\name}, respectively.  }}
    \label{fig:case}
    \vspace{-5mm}
\end{figure*} 


\begin{figure}[ht!]
	\centering
     \vspace{1mm}
    \includegraphics[width=1.0\linewidth]{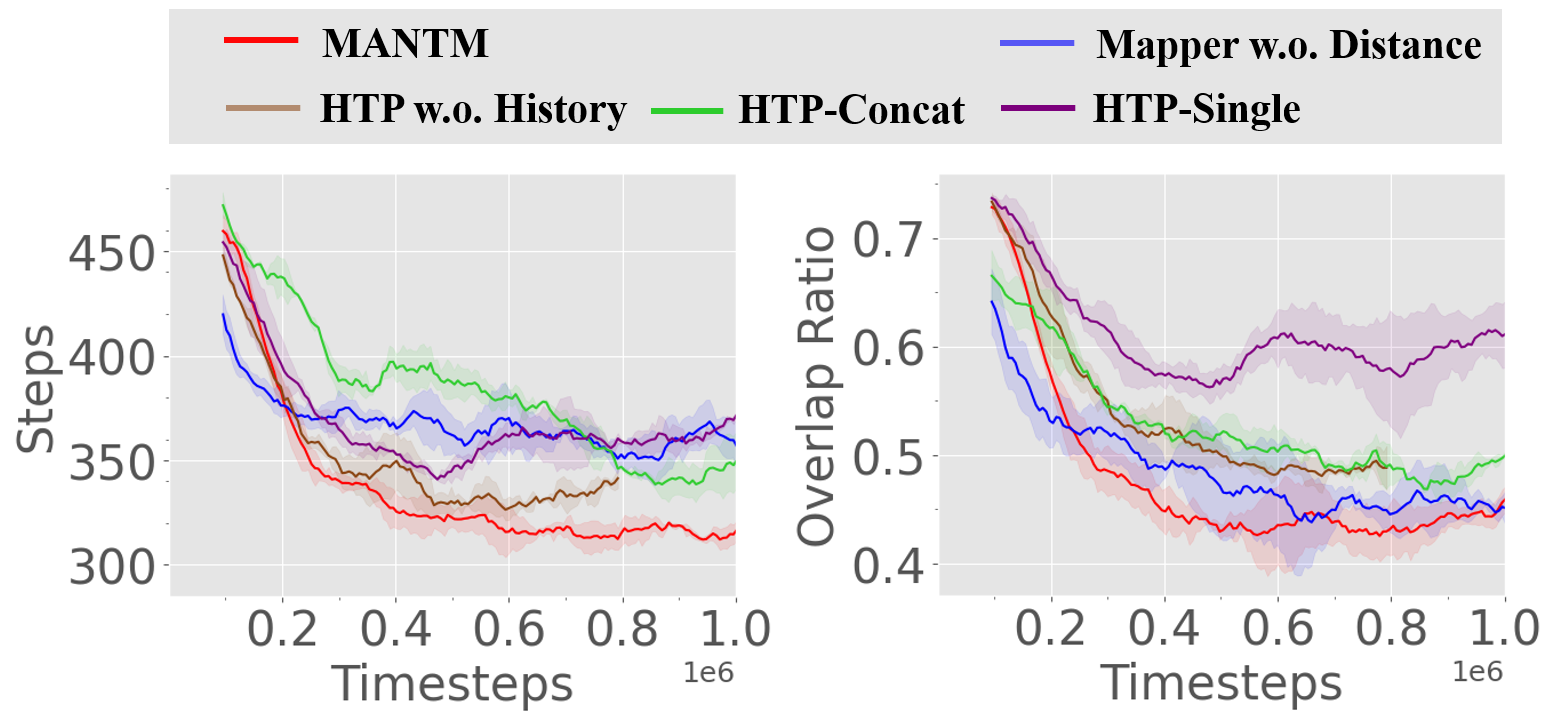}
    \vspace{-4mm}
	\centering \caption{Comparison between {\name} (red) and its variants. {\name} has the lowest \emph{Steps}
and \emph{Mutual Overlap}}
\label{fig:ab}
\vspace{-6mm}
\end{figure}
\subsection{Main Results}
\subsubsection{Evaluation Results}
The results in \cref{tab: gibson_results} show that {\name} outperforms all baselines with $N=3,4$ agents on unseen scenes on Gibson.
In both middle and large maps, {\name} attains the fewest \emph{Steps} and the highest \emph{Coverage} ratio. More concretely, MAANS is the best RL baseline since its transformer-based Spatial-TeamFormer captures the spatial relationship and team representation, while {\name} has 10.10\% and 7.63\% fewer \emph{Steps} than MAANS with $N=3,4$ agents, respectively. This indicates that ghost nodes, which are always located in unexplored areas, motivate agents to explore unseen regions. {\name} also excels in the \emph{Mutual Overlap} ratio among the RL solutions, suggesting that it better assigns global goals to agents in different unexplored directions.\looseness=-1

Among the planning-based baselines, the best competitor, Voronoi, achieves the lowest \emph{Mutual Overlap} ratio in the 3-agent setting, demonstrating that the Voronoi separates agents to reduce the overlapped area. However, {\name} is superior to Voronoi in the \emph{Steps} and the \emph{Coverage} ratio, reducing \emph{Steps} by 26.40\% and 31.57\% for $N=3,4$ agents, respectively.
This implies that although the Voronoi partition separates agents, it may hinder exploration efficiency by confining agents to stay in their respective areas. 



\subsubsection{Domain Generalization}
We also report the domain generalization performance in \cref{tab: hm3d_results}, where
all models trained on the Gibson dataset are evaluated in the HM3D domain.
The results indicate that {\name} performs best. Compared to {MAANS}, the best competitor, {\name} achieves 13.45\% and 11.04\% fewer \emph{Steps} with $N=3,4$ in super large scenes. This implies that the trained MAANS may overfit to the spatial arrangements in the training maps, resulting in suboptimal performance on unseen scenes in other domains due to variations in different metric maps.
In contrast, {\name} exploits the graph structure with abstract but essential information to better adapt to scenarios in an unseen domain. \looseness=-1

The planning-based methods fail to achieve an average \emph{Coverage} ratio of 90\% on super large maps. 
This indicates that the factors affecting exploration success are more complex on unseen super large maps. It is difficult for planning-based agents to take all factors into account in manual parameter tuning. As a result, agents may get stuck in the corner and fail to reach 90\% \emph{Coverage}.

\subsubsection{Case Study}
We present case studies of map construction in two different scenes to showcase the generalization of {\name}. Besides, to further demonstrate the cooperative exploration strategy of {\name}, we visualize the planning strategies of {\name} and the most competitive planning-based method, Voronoi.


In Fig.~\ref{fig:casea}, the graph structures in two different scenes appear to be generally congruent (i.e., the area in grey). This suggests that topological maps, which contain abstract but essential information, are less influenced by scene structures, endowing them with significant generalization capabilities. Conversely, the shape of metric maps depends on the layout of the scene. As a result, finding a (near) optimal exploration strategy for various metric maps is challenging.
In Fig.~\ref{fig:caseb}, the agent trajectories show that {\name} successfully allocates agents to different unexplored areas via the selected ghost nodes. Moreover, {\name} agents can temporarily revisit previously explored areas to reach unexplored areas in different directions, thereby increasing the final coverage. 
On the contrary, when the only path to the unexplored area belongs to the partition of a particular Voronoi agent (i.e., the area in blue), the other agent can only be constrained to its own partition (i.e., the area in green), resulting in inefficient exploration.

\vspace{-1mm}
\subsection{Ablation Study}
We report the training performance of several RL variants to investigate the importance of the Mapper and the {\planner}.

\begin{itemize}
\setlength{\parskip}{0pt} \setlength{\itemsep}{0pt plus 1pt}
 \item \textbf{Mapper w.o. Distance: }Without the help of the FMM distance, the Topological Mapper constructs the graph based solely on the similarity of the visual embeddings. 
  \item \textbf{{\planner} w.o. History: }We abandon the historical graphs, $\hat{G}$, and the Graph Fusion. The Main Node Selector only takes $G$ as input.
 \item \textbf{{\planner}-Single: }We abandon the Main Node Selector and only adopt the Ghost Node Selector to infer global goals.
 \item \textbf{{\planner}-Concat: }Before calculating $S_{g, re}$ in the Ghost Node Selector, we update ghost node features by concatenating them and the corresponding main node features. We then discard the multiplication in Equation~\ref{eq:multiply} and directly consider $S_{g, re}$ as the final matching score.
 
\end{itemize}

As shown in Fig.~\ref{fig:ab}, {\name} has the lowest \emph{Steps} and \emph{Mutual Overlap} ratio in $N$= 2 agents on Gibson. {\name} is superior to \emph{Mapper w.o. Distance} with over 10\% lower \emph{Steps}, suggesting that distance-based heuristics reduce incorrect connections between nodes and provide a more accurate graph for effective global planning.
Among all {\planner} variants, \emph{{\planner}-Single} has the highest \emph{Mutual Overlap} ratio and \emph{Steps}. This indicates that directly selecting ghost nodes as global goals may lead to a sub-optimal solution due to the large number of node candidates. The performance of \emph{{\planner} w.o. History} is worse in \emph{Mutual Overlap} ratio, suggesting that the lack of historical memory affects cooperation.
\emph{{\planner}-Concat} shows the lowest training convergence. The result expresses that the concatenation of the main and ghost node features prevents the {\planner} from better perceiving the relationship between these two types of nodes.

\section{Conclusion and Limitations}
We propose \emph{Multi-Agent Topological Neural Mapping} (\name), a multi-agent topological exploration framework, to improve exploration efficiency and generalization. In {\name}, the Topological Mapper constructs graphs via a visual encoder and distance-based heuristics. The RL-based Hierarchical Topological Planner (\planner) captures the relationships between agents and graph nodes to infer global goals. Experiments in Habitat demonstrate that {\name} outperforms planning-based baselines and RL variants in unseen scenes. However, there is a huge room for improvement in {\name}. For example, our method leverages metric maps to prune graphs, thus the precision of the metric map may influence the quality of the topological maps. Besides, we assume fully synchronized decision-making, which may fail in environments with communication latency. 

\bibliographystyle{IEEEtran}
\bibliography{IEEEfull}

\end{document}